\documentclass{article}

    \usepackage[preprint,nonatbib]{neurips_2024}

\usepackage[utf8]{inputenc} % allow utf-8 input
\usepackage[T1]{fontenc}    % use 8-bit T1 fonts
\usepackage{hyperref}       % hyperlinks
\usepackage{url}            % simple URL typesetting
\usepackage{booktabs}       % professional-quality tables
\usepackage{amsfonts}       % blackboard math symbols
\usepackage{nicefrac}       % compact symbols for 1/2, etc.
\usepackage{microtype}      % microtypography
\usepackage{xcolor}         % colors
\usepackage{graphicx}
\usepackage{wrapfig}
\usepackage{subfigure}
\usepackage{multirow}
\usepackage{amssymb}
\usepackage{pifont}
\usepackage{hyperref}
\usepackage{algorithm}
\usepackage{algorithmic}

\usepackage{amsmath}
\usepackage{amssymb}
\usepackage{mathtools}
\usepackage{amsthm}

\usepackage[capitalize,noabbrev]{cleveref}

\title{Relating-Up: Advancing Graph Neural Networks through Inter-Graph Relationships}

\author{%
    Qi Zou$^{1}$, Na Yu$^{1}$, Daoliang Zhang$^{1}$, Wei Zhang$^{1}$, Rui Gao$^{1}$\\
    School of control science and engineering, Shandong University, Jinan, China\\
    \texttt{\{qizou,yunacsw,zhangdaoliang\}@mail.sdu.edu.cn}, \\
    \texttt{\{zw,gaorui\}@sdu.edu.cn}
}

\begin{document}

\maketitle

\begin{abstract}
Graph Neural Networks (GNNs) have excelled in learning from graph-structured data, especially in understanding the relationships within a single graph, i.e., intra-graph relationships. Despite their successes, GNNs are limited by neglecting the context of relationships across graphs, i.e., inter-graph relationships. Recognizing the potential to extend this capability, we introduce Relating-Up, a plug-and-play module that enhances GNNs by exploiting inter-graph relationships. This module incorporates a relation-aware encoder and a feedback training strategy. The former enables GNNs to capture relationships across graphs, enriching relation-aware graph representation through collective context. The latter utilizes a feedback loop mechanism for the recursively refinement of these representations, leveraging insights from refining inter-graph dynamics to conduct feedback loop. The synergy between these two innovations results in a robust and versatile module. Relating-Up enhances the expressiveness of GNNs, enabling them to encapsulate a wider spectrum of graph relationships with greater precision. Our evaluations across 16 benchmark datasets demonstrate that integrating Relating-Up into GNN architectures substantially improves performance, positioning Relating-Up as a formidable choice for a broad spectrum of graph representation learning tasks.
\end{abstract}

\begin{wrapfigure}{r}{0.4\textwidth}
    \vskip -0.1in
    \centering
    \includegraphics[width=0.84\linewidth]{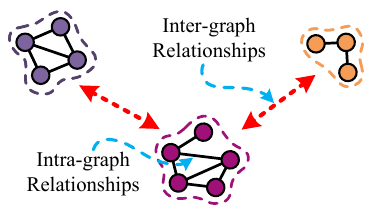}
    \caption{An illustration of inter-graph relationships. Inter-graph relationships can be similarities or differences in graph properties, structure, or other aspects. Such relationships advancing graph neural networks by capture the dynamics and interactions across graphs.}
    \label{fig1}
% \end{figure}
\end{wrapfigure}

\section{Introduction}

Graph Neural Networks (GNNs) have emerged as a specialized tool for graph representation learning, achieving remarkable success in node classification \cite{kipf2017semisupervised}, link predictions \cite{zhang_linkprediction}, as well as graph classification \cite{zhang2018end,errica2019fair,ijcai2019p601} and graph clustering \cite{ZahirniaSNL22,ijcai_0003LLW19} tasks, etc. The prevailing designs of modern GNNs employ message-passing strategies, excelling in capturing node relationships within a single graph by recursively aggregating information from neighbors to refine node representation. As depicted in Figure \ref{fig1}, despiting the effectiveness of this representation learning paradigm \cite{xu2018powerful,wijesinghe2022a}, existing GNN architectures is limited focusing to single graph, i.e., intra-graph relationships, the exploration of insightful relationships across graphs, i.e., inter-graph relationships, has neglected. For instance, consider the challenge in chemistry of distinguishing chiral molecules. While these molecules may seem similar in structure or atomic composition, their unique interactions in ligand or enzyme-catalyzed reactions highlight distinct differences. Thus, there is a need to move beyond single graph and devise method to extract relationships spanning graphs.

\begin{figure*}[!t]
    \vskip -0.1in
    \centering
    \includegraphics[width=0.85\textwidth]{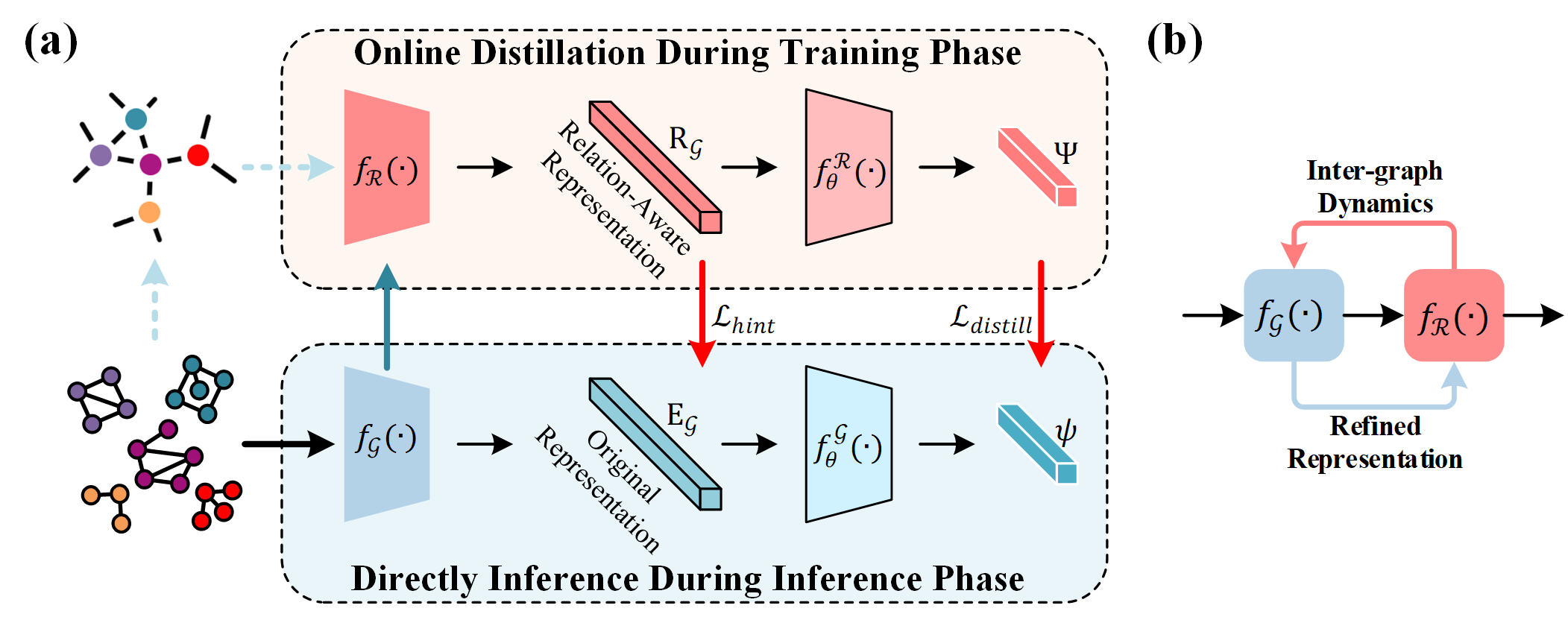}
    \caption{Framework of Relating-Up. \textbf{(a)} The overall architecture of Relating-Up. Both graph encoder and relation encoder are jointly optimized from scratch. During the training phase, the relation encoder $f_\mathcal{R}$ dynamically interprets inter-graph relationships to iteratively refines graph encoder $f_\mathcal{G}$ through a feedback training strategy. During the inference phase, the refined graph encoder $f_\mathcal{G}$ is directly deployed for predictions. \textbf{(b)} Information flow in a feedback loop between graph encoder and relational encoder. This cyclical process allows for an evolving understanding of dynamics across graphs, where each iteration contributes to a more profound comprehension of the inter-graph relationships.}
    \label{fig2}
    \vskip -0.05in
\end{figure*}

Capturing and utilizing inter-graph relationships remains an challenging problem. A straightforward approach relies on heuristical design, utilizing data augmentation \cite{zhao2023graph,Gasteiger_NIPS_2019} to generate variation graphs from a vicinity around training graphs to enriching the training distribution. Other methods, such as contrastive learning \cite{NEURIPS2020_3fe23034}, by emphasizing intra-consistency and inter-differences, aspire to group similar entities while distancing diverse entities. However, these methods remain inherently bound to the paradigm of intra-graph focus, limiting their ability in explicitly unveiling inter-graph relationships.

In this paper, we introduce Relating-Up, a novel module designed to enhance existing GNN architectures by dynamically utilizing inter-graph relationships. Unlike previous efforts that seek performance improvements through complex and specialized techniques, Relating-Up offers a straightforward yet effective approach to enhance GNNs by tapping into underutilized context of relationships between graphs. Specifically, Relating-Up incorporates a feedback loop mechanism that enriches graph representation learning by bridging conventional graph representation with an innovative relation-aware representation. Through continuous mutual refinement, these representation spaces significantly improve the expressiveness and accuracy of graph representations, deepening our understanding of inter-graph dynamics. Our assessments on various benchmark datasets reveal that incorporating the Relating-Up module into GNN architectures significantly improves their effectiveness. This positions Relating-Up as a strong option for a wide range of tasks in graph representation learning.

In summary, our contributions are highlighted as follows:
\begin{enumerate}
    \item Relating-Up introduces the concept of inter-graph dynamics into GNN architectures, enabling these models to understand and leverage the contextual information present across graphs.
    \item We propose an feedback training strategy that iteratively refines graph representations by leveraging insights from inter-graph dynamics, ensuring that the learning process is not static but evolves dynamically, leading to more robust and generalizable graph models.
    \item Relating-Up is designed to be a versatile, plug-and-play module that can be integrated seamlessly with existing GNN architectures, ensuring compatibility across a wide range of models and expanding the applicability of GNNs.
\end{enumerate}

\section{Related Work}

\subsection{Graph Neural Networks}

GNNs attempt to exploit the relational information in real-world, thereby enhancing various downstream fields, such as molecules\cite{ijcai2023p760}, recommendation\cite{CSUR_2022_WU}, and social networks\cite{Li_2023_TPAMI_Semi,Wu_Lian_Xu_Wu_Chen_2020}. Pioneering architectures such as GCNs \cite{kipf2017semisupervised}, GraphSAGE \cite{HamiltonYL17}, GATs \cite{veličković2018graph}, GIN \cite{xu2018powerful} have ingeniously introduced the convolution operation to graphs. The foundational philosophy behind these designs is the recursive aggregation of information from neighboring nodes \cite{xu2018representation,xu2018powerful}, epitomizing the classic GNN paradigm that emphasizes neighborhood aggregation to discern local intra-graph relationships. In the specific field of graph classification, pooling is usually used to obtain the representation of the entire graph \cite{duvenaud2015convolutional,zhang2018end,pmlr_v97_lee19c}. These methods form efficient representation of the entire graph by shrinking graph into different scales. Although many variants of GNNs with different neighborhood aggregation and graph-level pooling schemes have been proposed, their main focus remains on intra-graph, mapping the representation of a graph by summarizing node information enhanced through intra-graph structure, inadvertently neglecting inter-graph relations.

\subsection{Graph Data Augmentation}
Despite the exceptional prowess of GNNs in theoretical settings, their practical efficacy on real-world graphs often remains precarious \cite{pmlr_v139_balcilar21a}. To address the overreliance on labeled data and intrinsic oversomoothing issue, graph data augmentation has emerged as a pivotal technique to enhance the robustness and generalization of GNNs\cite{Ding_SIGKDD_2022}. Graph data augmentation typically involves creating synthetic graphs \cite{Gasteiger_NIPS_2019} or modifying existing ones \cite{NEURIPS2021_b8b2926b} to reflect potential relationships within a transformation neighborhood, encouraging the model to learn more generalizable patterns. While these approaches internally exploit the relationships among variants graphs within a transformation neighborhood \cite{Wang_WWW21_2021}, they do not explicitly facilitate learning inter-graph relationships, which is a critical aspect of comprehensively understanding graph data. 

\subsection{Graph Contrastive Learning}
Graph contrastive learning is a burgeoning area in graph representation learning, focused on extracting meaningful patterns from graphs by emphasizing contrasts between predefined graph relationships. Graph contrastive learning aim to maximizing the similarity between representations of similar graphs while minimizing it for dissimilar ones \cite{NEURIPS2020_3fe23034,Hassani_2020_ICML}. However, this approach encounters a critical limitation in its inability to adaptively handle inter-graph relationships. The effectiveness of graph contrastive learning is contingent upon predefined priors knowledge, which limits its scope of application. It often struggles to autonomously uncover hidden or implicit relationships across graphs, thereby overlooking insightful connections.

\subsection{Advancement in Inter-graph Relationships}
Recent advancements have utilized semi-supervised learning and class-aware refinements for graph classification, such as SEAL \cite{Li_2023_TPAMI_Semi} and CARE \cite{xu2022classaware}. Diverging from these methods, our proposed module introduces a dynamic mechanism that continuously evolves graph representations by leveraging inter-graph relationships, enhancing robustness and contextual awareness. Unlike SEAL, Relating-Up adaptively uncovers complex relationships across graphs. Furthermore, it extends the scope of graph representation by exploring the varied structural and feature-based relationships across graphs, beyond traditional class labels.

Overall, our proposed Relating-Up addresses these shortcomings by inherently leveraging inter-graph relationships. It adaptively and dynamically learns the complex relationships across graphs, offering a more comprehensive and insightful perspective of graph data.

\section{Preliminaries}
In this section, we introduce the fundamental concepts and notations that underpin our work, providing a groundwork for understanding our proposed Relating-Up module.

\subsection{Notations}

In the context of graph representation learning tasks, we consider a graph denoted as $\mathcal{G} = (\mathcal{V}, \mathcal{E})$, where $\mathcal{V}$ represents the set of nodes and $\mathcal{E}$ signifies the set of edges. Each node $v \in \mathcal{V}$ is characterized by a feature vector $\mathbf{x}_v \in \mathbb{R}^{d}$, and the total number of nodes is given by $\vert\mathcal{V}\vert = n$. The edges, $\mathcal{E}$, play a pivotal role as they define the intra-graph relationships between the nodes.

The primary objective in graph representation learning tasks is to learn a function $f_\mathcal{G}: \mathcal{G} \rightarrow \mathbf{E}_{\mathcal{G}}$. The $f_\mathcal{G}$ is responsible for encoding the graph $\mathcal{G}$ into an informative representation space $\mathbf{E}_{\mathcal{G}} \in \mathbb{R}^{d_g}$, which is instrumental for various downstream tasks. Specifically, for graph classification tasks, this involves combining a classification head $f_{\theta}: \mathbf{E}_{\mathcal{G}} \rightarrow y$ trained on a set of graphs $\mathbb{G}=\{(\mathcal{G}, y)\}$ to facilitate accurate classification. 

\subsection{Graph Encoder}
To effectively capture the intricate relationships among nodes, GNN architectures recursively updating node representations according to the intra-graph relationships among neighboring nodes using a message passing strategy, formalized as follows:
\begin{align}
    & \mathbf{m}_{v}^{l+1} = \text{Aggregate}^{l+1}(\{\{\mathbf{h}_{u}^{l} : u \in \mathcal{N}(v)\}\}), \nonumber \\
    & \mathbf{h}_{v}^{l+1} = \text{Combine}^{l+1}(\mathbf{h}_{v}^{l}, \mathbf{m}_{v}^{l+1}),
\end{align}
where $\mathbf{h}_{v}^{l+1}$ represents the node representation of node $v$ after the $l-\text{th}$ layer, with $\mathbf{h}_{v}^{0}$ initialized as $\mathbf{x}_v$. The message-passing strategy for each GNN layer simplifies to $\mathbf{h}_{v}^{l+1} = \text{GNN}^{l+1}(\mathbf{h}_{v}^{l}, \mathcal{N}(v))$, where $\mathcal{N}(v)$ denotes the neighborhood of node $v$, representing intra-graph relationships. The functions $\text{Aggregate}(\cdot)$ and $\text{Combine}(\cdot)$ are parametric functions designed to aggregate neighborhood information and combine them to update node representation.

By stacking multiple message-passing layers, a set of node representations is produced. The overall graph representation, $\mathbf{E}_\mathcal{G} \in \mathbb{R}^{d_{g}}$, is then obtained through a $\text{Readout}(\cdot)$ function as defined by:
\begin{equation}
    \mathbf{E}_g = \text{Readout}(\{\{\mathbf{h}_{v}^{k} : v \in \mathcal{V}\}\}),
\end{equation}
where $\text{Readout}(\cdot)$ is a permutation invariance function to capture the structural nuances of the graph.

\subsection{Graph Classification}
For a given graph $\mathcal{G}\in \mathbb{G}$, the graph encoder $f_\mathcal{G}$ transform individual graph structures and proprieties into fixed-dimensional representations $\mathbf{E}_\mathcal{G}$. Subsequently, $\mathbf{E}_\mathcal{G}$ is fed into the classification head, denoted as $f_{\theta}$, to obtain class probabilities. The loss function is typically defined as the CrossEntropy of the predictions over the labels:
\begin{align}
    \mathcal{L}=\text{CrossEntropy}(\hat{y},y)=-\sum_{\mathcal{G} \in \mathbb{G}} \sum_{c \in \text{C}} y_{gc} \log(\hat{y}_{gc}),
\end{align}
where $\hat{y}_{gc}=\text{softmax}(f_{\theta}(\mathbf{E}_\mathcal{G}))$ denotes the predicted probability for graph $\mathcal{G}$ belonging to class $c$, and $y_{gc}$ is the ground truth label.

\section{The Proposed Methods}

In this section, we describe the proposed Relating-Up module. As shown in in Figure \ref{fig2}, Relating-Up consisting of three key components: a graph encoder, a relation encoder, and a feedback training strategy. By focusing on the extension of GNNs rather than introducing an entirely new framework, Relating-Up respects the foundational strengths of GNNs while addressing their limitations in considering inter-graph relationships.

\subsection{Graph Encoder} 
In Relating-Up, we implement graph encoder $f_{\mathcal{G}}$ by stacking multiple GNN layers, which extracts a fixed-dimensional representation $\mathbf{E}_{g}$ for each graph $\mathcal{G} \in \mathbb{G}$. Notably, Relating-Up is designed for integrate seamlessly with various GNN backbones without altering their fundamental operations.

\subsection{Relation Encoder}
Upon deriving the graph representations, the relation encoder, denoted as $f_\mathcal{R}$, dynamically interprets the relationships between these graphs. Specifically, it processes a set of graph representations, symbolized as $\mathbf{E} \in \mathbb{R}^{|\mathbb{G}|\times d_g}$. Each layer of $f_\mathcal{R}$ consists of a multi-head self-attention (MSA) mechanism and a feed-forward network (FFN) block, each followed by a layer normalization (LN). The MSA, pivotal to the encoder, dynamically computing inter-graph relationships as:
\begin{equation}
    \text{Attention}(\mathbf{Q},\mathbf{K},\mathbf{V})=\text{softmax}(\frac{\mathbf{Q}\mathbf{K}^T}{\sqrt{d_k}})\mathbf{V},
\end{equation}
where $\mathbf{Q} \in \mathbb{R}^{|\mathbb{B}|\times d_q}$, $\mathbf{K} \in \mathbb{R}^{|\mathbb{B}|\times d_k}$, $\mathbf{V} \in \mathbb{R}^{|\mathbb{B}|\times d_v}$ represent queries, keys, and values, respectively, derived from the graph representations. $|\mathbb{B}|$ is the batch size during the training phase. Each head in the MSA captures unique relationship patterns, with their outputs concatenated and linearly transformed for comprehensive relationship encoding:

\begin{align}
    \text{MSA}(\mathbf{Q},\mathbf{K},\mathbf{V})= [\mathbf{O}_1,\mathbf{O}_2,...,\mathbf{O}_h]\mathbf{W}^O,  \nonumber \\
    \mathbf{O}_i =\text{Attention}_i(\mathbf{Q}{\mathbf{W}}^Q_i,\mathbf{K}{\mathbf{W}}^K_i,\mathbf{V}{\mathbf{W}}^V_i),
\end{align}
where the operations for $h$ parallel projections are parameter $\mathbf{W}^Q_i\in \mathbb{R}^{|\mathbb{B}| \times d_k}$, $\mathbf{W}^K_i\in \mathbb{R}^{|\mathbb{B}| \times d_k}$, and $\mathbf{W}^V_i\in \mathbb{R}^{|\mathbb{B}| \times d_v}$, where $h$ stands for the number of attention heads. The output projection matrix is $\mathbf{W}^O \in \mathbb{R}^{h d_v \times d_{g}}$. Then, the inter-graph relation-aware representation at $l$-th relation layer is formulated as:
\begin{equation}
    \mathbf{R} = f_\mathcal{R}(\{\{ \mathbf{E}_g | g \in \mathbb{G}  \}\}),
\end{equation}
where $\mathbf{R}_{\mathcal{G}} \in \mathbb{R}^{d_g}$ denotes the refined relation-aware representations, now enriched with inter-graph relationship information. This approach allows Relating-Up to dynamically and adaptively learns the nuanced relationships across graphs, which is crucial for tasks that require a comparative understanding of the inter-graphs relationships.

\subsection{Feedback Training Strategy}

We introduce an innovative feedback training strategy as a pivotal component of the Relating-Up module. Algorithm \ref{alg:feedback} reports the pseudo-code of the feedback training strategy. This strategy is ingeniously designed to dynamically utilize the potential inter-graph relationships. Specifically, as shown in Figure \ref{fig2} (b), we utilize a feedback loop mechanism to progressively refine graph representations. Mathematically, this strategy involves minimizing the discrepancy between the original graph representation space and relation-aware representation space,ensuring effective integration of insights from the relation-aware space into the original graph representation. To achieve this, we construct the feedback loop mechanism by two classifier $f^{\mathcal{G}}_{\theta}$ and $f^{\mathcal{R}}_{\theta}$ after graph encoder and relation encoder, respectively. The relation-aware insights from relation encoder are then distilled and conveyed to graph encoder to established a dynamic feedback loop:
\begin{equation}
    \mathcal{L}=\sum_{\mathcal{G} \in \mathbb{G}} \alpha \mathcal{L}_{class}(\mathcal{G}) + (1 - \alpha) \mathcal{L}_{distill}(\mathcal{G}) + \beta \mathcal{L}_{hint}(\mathcal{G}),
\end{equation}
where $\mathcal{L}_{class}(\mathcal{G})$ is the CrossEntropy loss for the labeled graph entities from the $f^{\mathcal{G}}_{\theta}$ and $f^{\mathcal{R}}_{\theta}$. The $\mathcal{L}_{distill}(\mathcal{G})$ is formulated as:
\begin{equation}
    \mathcal{L}_{distill}(\mathcal{G})=D_{kl}(\psi, \Psi) 
\end{equation}
where $D_{kl}(\cdot,\cdot)$ is the Kullback-Leibler divergence,  $D_{kl}(p,q)=\sum_j p_j log \frac{p_j}{q_j}$. $\psi = \text{softmax}(f^{\mathcal{G}}_{\theta}(\mathbf{E_\mathcal{G}})/T)$ and $\Psi = \text{softmax}(f^{\mathcal{R}}_{\theta}(\mathbf{R_\mathcal{G}})/T)$ are $f^{\mathcal{G}}_{\theta}$ and $f^{\mathcal{R}}_{\theta}$ predicted probability distribution, respectively. $T$ is the temperature of distillation with a larger $T$ makes the probability distribution softer. The relation hints is defined as the inexplicit knowledge in relation-aware representation to guide $f_{\mathcal{G}}$ to glimpse the global landscape of $f_{\mathcal{R}}$ during the training period. It works by decreasing the L2 distance between $\mathbf{E}_{\mathcal{G}}$ and $\mathbf{R}_{\mathcal{G}}$:
\begin{equation}
    \mathcal{L}_{hint}(\mathcal{G})=\parallel\mathbf{R}_{\mathcal{G}} - \mathbf{E}_{\mathcal{G}}\parallel_2^2.
\end{equation}

\begin{table}[h]
    \caption{Graph classification in terms of accuracy (\%) with standard deviation for Relating-Up combined with different GNN backbones. The best results are \textbf{boldfaced}. The paired t-test between Relating-Up and original backbones at significance level of 10\% are \underline{underlined}.}
    \vskip -0.05in
    \label{tab_backbone}
    \centering
    \scalebox{0.65}{
    \begin{tabular}{lcccccccc}
        \toprule
        \multicolumn{1}{@{}c@{}}{\multirow{2.5}{*}{Dataset}} 
        & \multicolumn{2}{@{}c@{}}{GraphSAGE}
        & \multicolumn{2}{@{}c@{}}{GCN} 
        & \multicolumn{2}{@{}c@{}}{GAT}
        & \multicolumn{2}{@{}c@{}}{GIN} \\
          \cmidrule(l){2-3} 
          \cmidrule(l){4-5}   
          \cmidrule(l){6-7}
          \cmidrule(l){8-9}
          
          & \multicolumn{1}{@{}c@{}}{Original} & \multicolumn{1}{@{}c@{}}{Relating-Up} 
          & \multicolumn{1}{@{}c@{}}{Original} & \multicolumn{1}{@{}c@{}}{Relating-Up} 
          & \multicolumn{1}{@{}c@{}}{Original} & \multicolumn{1}{@{}c@{}}{Relating-Up}
          & \multicolumn{1}{@{}c@{}}{Original} & \multicolumn{1}{@{}c@{}}{Relating-Up}\\
          
        \midrule
        MUTAG                    
                & 83.04±11.20 & \textbf{85.67±7.45}
                & 84.09±10.23 & \textbf{86.20±7.87} 
                & 81.93±11.90 & \textbf{85.09±9.40}
                & 85.70±9.08  & \textbf{86.70±7.55} \\
                                 
        PTC-FM
                & 61.89±4.24  & \textbf{64.48±5.83}
                & 60.73±3.05  & \underline{\textbf{64.18±4.63}}
                & 60.18±7.02  & \underline{\textbf{63.91±5.51}}
                & 57.89±4.13  & \underline{\textbf{64.47±4.06}} \\
                                 
        PTC-MM
                & 61.61±6.96  & \underline{\textbf{66.40±6.30}}
                & 65.20±5.21  & \underline{\textbf{67.30±4.90}}
                & 64.59±5.16  & \textbf{65.77±5.63}
                & 62.53±7.91  & \underline{\textbf{66.38±4.10}} \\
                                 
        PTC-FR                  
                & 66.11±2.77  & \textbf{67.24±4.45}
                & 65.81±2.57  & \textbf{66.94±4.13}
                & 65.25±2.41  & \underline{\textbf{67.81±1.84}}
                & 66.07±4.28  & \textbf{68.93±5.56} \\
                                 
        PTC-MR                  
                & 59.85±7.45  & \textbf{61.04±8.26}
                & 56.65±6.08  & \underline{\textbf{61.55±9.24}}
                & 57.07±8.31  & \underline{\textbf{60.99±6.94}}
                & 57.56±6.29  & \underline{\textbf{62.46±5.80}} \\
                                 
        COX2                     
                & 81.60±4.24  & \textbf{83.70±5.36} 
                & 81.37±4.48  & \textbf{83.07±4.66} 
                & 77.94±1.02  & \underline{\textbf{79.23±2.03}}
                & 83.50±3.63  & \textbf{83.72±4.10} \\
                                 
        COX2-MD                 
                & 60.43±10.44 & \underline{\textbf{66.40±8.61}}
                & 57.48±7.54  & \underline{\textbf{66.73±7.83}}
                & 65.08±9.01  & \textbf{67.04±9.03}
                & 60.46±9.03  & \textbf{67.63±8.40} \\
                                 
        PROTEINS                 
                & 72.68±2.65  & \underline{\textbf{74.66±3.25}}
                & 73.49±3.71  & \textbf{74.84±3.78}    
                & 73.22±2.09  & \underline{\textbf{74.62±3.46}}
                & 73.67±3.57  & \underline{\textbf{75.91±4.41}} \\
                                 
        DD                      
                & 77.93±3.08  & \textbf{78.27±3.06}             
                & 76.66±3.50  & \textbf{77.84±3.30}    
                & 76.18±3.70  & \underline{\textbf{78.27±2.61}}   
                & 76.57±2.55  & \underline{\textbf{79.12±3.84}} \\
                                         
        NCI1                     
                & 76.41±2.13  & \textbf{76.73±2.16}             
                & 76.33±2.53  & \textbf{77.19±1.65}             
                & 67.01±2.30  & \textbf{67.88±1.56}
                & 79.39±1.42  & \textbf{80.66±2.65} \\
                                         
        NCI109       
                & 75.41±2.98  & \textbf{75.70±2.22}            
                & 75.02±2.35  & \underline{\textbf{76.75±2.17}}
                & \textbf{66.97±2.89}  & 66.42±2.41
                & 78.41±2.45  & \underline{\textbf{80.63±2.10}} \\
                                 
        ENZYMES                 
                & 36.83±6.21  & \underline{\textbf{40.83±4.23}}
                & 38.33±6.15  & \textbf{40.83±3.59}
                & 28.67±4.70  & \textbf{30.33±4.33}
                & 40.00±5.77  & \underline{\textbf{44.50±6.06}}\\
                                 
        \midrule
        IMDB-B             
                & 72.00±2.57  & \textbf{73.10±2.47}             
                & 73.00±3.41  & \textbf{74.50±3.35}            
                & 71.70±2.37  & \underline{\textbf{73.50±2.55}}
                & 73.70±2.69  & \underline{\textbf{75.80±3.49}} \\
                                 
        IMDB-M              
                & 50.93±3.53  & \textbf{51.53±4.30}             
                & 50.87±3.64  & \textbf{52.13±3.51}             
                & 50.80±3.62  & \textbf{51.53±2.88}
                & 49.40±3.39  & \underline{\textbf{52.00±3.35}} \\
                                 
        REDDIT-B          
                & 74.80±1.91  & \textbf{75.01±1.83}            
                & 74.75±3.30  & \underline{\textbf{78.05±3.12}}
                & 74.75±1.57  & \textbf{75.49±1.78}
                & 73.70±3.71  & \underline{\textbf{78.90±2.77}} \\
                                 
        REDDIT-M5          
                & 34.65±1.70  & \textbf{35.23±0.80}             
                & 36.89±1.88  & \underline{\textbf{39.65±1.11}} 
                & \textbf{33.83±1.11}  & 33.67±1.66
                & 51.23±2.71  & \underline{\textbf{54.11±2.10}} \\
        \bottomrule
    \end{tabular}}
    \vskip -0.1in
\end{table}

By embedding a feedback strategy that iteratively refines graph representations with the relation-aware insights, this strategy significantly enhances the capability of GNNs to understand and utilize the rich and complex tapestry of inter-graph relationships.

\subsection{Inference}

During training phase, the relation encoder and relation-aware classifier actively enhance the graph encoder with deep insights into inter-graph relationships. This process embeds a rich understanding of these relationships directly into the graph representations, ensuring that the encoder captures not only intra-graph features but also the nuanced dynamics between different graphs. As a result, during inference phase, the reliance shifts solely to this enriched graph encoder and original graph representation classifier for direct classification. This shift is underpinned by the premise that the relational insights, once integrated into the graph representations during training, remain inherently effective for classification tasks. The elimination of the relation encoder and relation-aware classifier during inference significantly streamlines the process, optimizing for speed and computational efficiency while maintaining the robustness and accuracy imparted during training.

\section{Experiments}

\begin{wraptable}{r}{0.48\textwidth}
    \vskip -0.24in
    \caption{Graph classification in terms of ROC-AUC (\%) with standard deviation on OGBG-MOLHIV dataset. The best results are \textbf{boldfaced}.}
    \vskip 0.12in
    \label{tab:ogb}
    \centering
    \scalebox{0.75}{
    \begin{tabular}{llcc}
        \toprule
        Backbone & Method & Validation & Test \\ 
        \midrule
        \multirow{2}{*}{GCN} & Original    & 81.23±1.14 & 76.09±0.45 \\
                             & Relating-Up & \textbf{85.53±0.44} & \textbf{77.84±0.81} \\
        \midrule
        \multirow{2}{*}{GCN-Virtual} & Original    & 83.98±1.00 & 74.90±0.86 \\
                                     & Relating-Up & \textbf{85.87±0.90} & \textbf{76.96±0.76} \\
        \midrule
        \midrule
        \multirow{2}{*}{GIN} & Original    & 81.16±0.77 & 74.58±1.13 \\
                             & Relating-Up & \textbf{82.87±0.42} & \textbf{75.48±1.09} \\
        \midrule
        \multirow{2}{*}{GIN-Virtual} & Original    & \textbf{84.70±0.54} & 76.65±2.11 \\
                                     & Relating-Up & 82.56±1.21 & \textbf{78.87±1.16} \\
        \bottomrule
    \end{tabular}}
    \vskip -0.12in
\end{wraptable}

In this section, we conduct comprehensive experiments to assess the effectiveness of the proposed method, which includes comparison experiments, empirical exploration of the proposed method, and ablation experiments. More exploratory experimental results are presented in Appendix. Our code is available at \url{https://anonymous.4open.science/r/RelatingUp-Q417}. 

\textbf{Dataset.} In our experiments, we use a diverse collection of 17 graph classification benchmark datasets, covering datasets from the bioinformatics domains to the social domains. Specifically, this collection includes 16 datasets collected from the TUDataset \cite{KKMMN2016} and the OGBG-MOLHIV dataset collected from the Open Graph Benchmark (OGB) \cite{hu2021open}. Detailed statistics and properties of these datasets are presented in Table \ref{tabs1}.

\textbf{Model.} The proposed Relating-Up module is a plug-and-play designed for seamless integration with existing graph representation learning architectures. We evaluate the effectiveness of Relating-Up with (1) different GNN backbones, including GCN \cite{kipf2017semisupervised}, GraphSAGE \cite{HamiltonYL17}, and GIN \cite{xu2018powerful}; (2) various graph pooling methods include global pooling, DiffPool, global attention pooling, GAPool, and self-attention graph pooling, SAPool \cite{pmlr_v97_lee19c,knyazev_2019_NeurIPS_understanding}; (3) other graph representation learning paradigms based on graph augmentation and contrastive learning that required explicit definition of graph relationships, e.g., graph augmentation, mixup for graph classification \cite{Wang_WWW21_2021}, GraphCL \cite{NEURIPS2020_3fe23034}, MVGRL\cite{Hassani_2020_ICML}.

\textbf{Implement Details.} 
To ensure fair comparisons, we standardized hyperparameters across various GNN based architectures. Specifically, each model was configured with five GNN layers with each comprising 128 hidden units. We employed a mini-batch size of 128 and a dropout ratio of 0.5. For optimization, the Adam optimizer was employed, starting with a 0.01 learning rate, halved every 50 epochs. 

\begin{table}[t]
    \caption{Graph classification in terms of accuracy (\%) with standard deviation for Relating-Up combined with different pooling operations. The best results are \textbf{boldfaced}.}
    \vskip -0.03in
    \label{tab_pooling}
    \begin{center}
    \scalebox{0.85}{
    \begin{tabular}{lllllll}
        \toprule
        \multicolumn{1}{@{}c@{}}{\multirow{2.5}{*}{Dataset}} 
        & \multicolumn{2}{@{}c@{}}{DiffPool} & \multicolumn{2}{@{}c@{}}{GAPool} & \multicolumn{2}{@{}c@{}}{SAGPool} \\
        \cmidrule(l){2-3}                      \cmidrule(l){4-5}                   \cmidrule(l){6-7}
        & \multicolumn{1}{@{}c@{}}{Original} & \multicolumn{1}{@{}c@{}}{Relating-Up} 
        & \multicolumn{1}{@{}c@{}}{Original} & \multicolumn{1}{@{}c@{}}{Relating-Up} 
        & \multicolumn{1}{@{}c@{}}{Original} & \multicolumn{1}{@{}c@{}}{Relating-Up} \\
        
        \midrule
        MUTAG     
            & 76.54±7.00 & \textbf{80.61±4.45} 
            & 76.61±8.97 & \textbf{78.22±8.25}
            & 84.04±6.29 & \textbf{84.53±8.73} \\
        
        PTC-FM   
            & 55.01±5.17 & \textbf{59.09±3.54} 
            & 63.05±5.09 & \textbf{65.04±6.10}
            & 61.63±8.00 & \textbf{64.76±4.37} \\
        
        PTC-MM   
            & 58.03±3.93 & \textbf{61.24±1.96}
            & 65.20±5.18 & \textbf{68.16±5.24} 
            & 64.03±6.97 & \textbf{68.46±6.19} \\
       
        PTC-FR   
            & 61.75±6.11 & \textbf{63.93±2.77}
            & 65.23±3.88 & \textbf{67.80±2.61} 
            & 65.50±6.67 & \textbf{68.37±3.73} \\
       
        PTC-MR   
            & 54.70±2.80 & \textbf{55.03±2.71} 
            & 58.12±3.95 & \textbf{61.05±6.61}
            & 56.38±3.82 & \textbf{60.16±3.82} \\
  
        COX2      
            & 77.20±2.71 & \textbf{78.52±1.88}
            & 82.86±3.06 & \textbf{83.26±2.38}
            & 80.30±4.05 & \textbf{83.29±4.25} \\
   
        COX2-MD  
            & 53.21±3.44 & \textbf{57.91±2.87}
            & 58.10±6.01 & \textbf{65.70±6.57} 
            & 62.10±9.21 & \textbf{66.37±8.49} \\
     
        PROTEINS  
            & 66.86±2.98 & \textbf{71.34±2.19}
            & 70.43±3.19 & \textbf{73.03±4.11} 
            & 73.84±4.10 & \textbf{75.91±3.70} \\
      
        DD        
            & 72.32±1.93 & \textbf{74.62±1.32}
            & 73.51±4.09 & \textbf{74.15±2.97} 
            & 75.46±2.55 & \textbf{78.02±2.02} \\
       
        NCI1      
            & 69.77±1.17 & \textbf{71.02±0.77}
            & 74.53±2.10 & \textbf{74.86±3.42   } 
            & 73.65±3.44 & \textbf{75.75±2.70} \\
        
        NCI109   
            & 67.29±2.02 & \textbf{69.43±1.34}
            & 73.42±2.11 & \textbf{73.75±2.70}
            & 74.10±2.08 & \textbf{75.83±2.12} \\
                                         
        ENZYMES                 
            & 30.80±4.34 & \textbf{32.63±2.73}
            & 36.83±6.21 & \textbf{37.33±5.28}
            & 34.50±6.01 & \textbf{38.83±5.78} \\
        \midrule
        IMDB-BINARY   
            & 64.80±3.95 & \textbf{66.65±1.54} 
            & 71.10±3.11 & \textbf{74.70±2.65}
            & 72.60±3.72 & \textbf{74.30±2.79} \\
        
        IMDB-MULTI   
            & 42.53±2.53 & \textbf{45.73±2.05} 
            & 50.53±3.05 & \textbf{52.87±3.17}
            & 49.07±3.13 & \textbf{52.00±3.40} \\
       
        REDDIT-BINARY  
            & 75.16±1.20 & \textbf{76.88±2.61} 
            & 74.80±1.91 & \textbf{75.44±1.96}
            & 89.02±2.49 & \textbf{89.40±2.54}\\
      
        REDDIT-MULTI-5K
            & 30.71±1.31 & \textbf{32.75±1.95} 
            & 34.65±1.70 & \textbf{34.83±1.73} 
            & 50.73±1.85 & \textbf{52.09±3.03} \\
            
        \bottomrule
    \end{tabular}}
    \end{center}
    \vskip -0.15in
\end{table}

\begin{figure}[!t]
    \centering
    \includegraphics[width=0.75\textwidth]{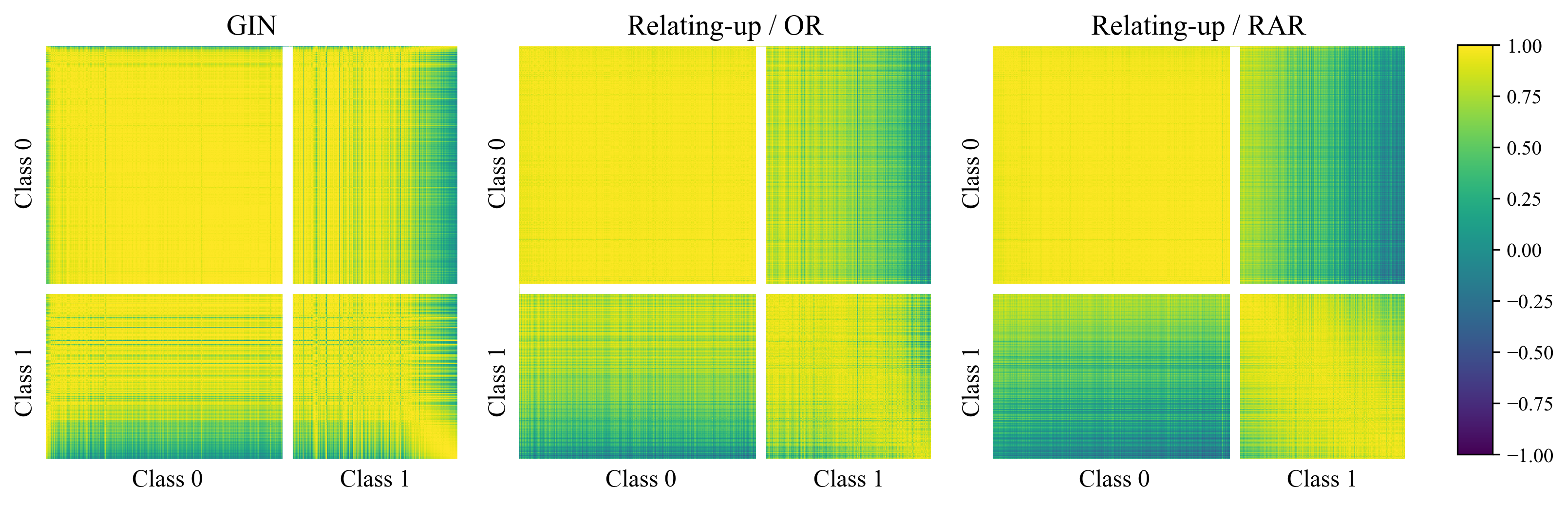}
    \caption{The graph embedding similarity of graph representation of DD dataset extracted from GNN backbone (GIN) and Relating-Up variants. \textbf{OR}: original representation, \textbf{RAR}: relation-aware representation.}
    \label{fig_heatmap}
    \vskip -0.15in
\end{figure}

\begin{figure}[!t]
    \centering
    \includegraphics[width=0.85\textwidth]{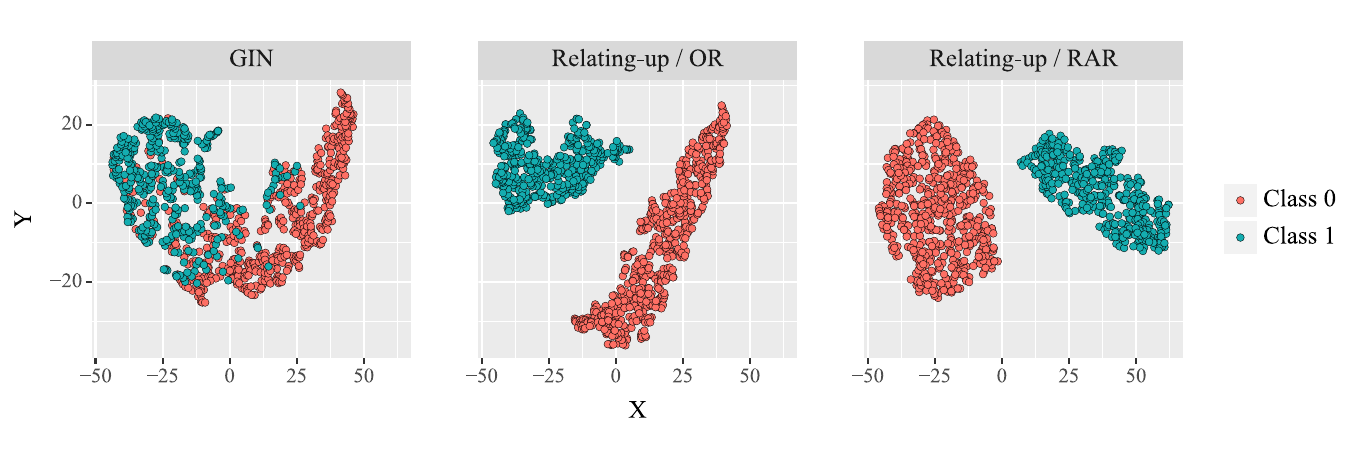}
    \vskip -0.1in
    \caption{The t-SNE two-dimensional embedding of graph representation of DD dataset extracted from GNN backbone (GIN) and Relating-Up variants. \textbf{OR}: original representation, \textbf{RAR}: relation-aware representation.}
    \label{fig_tsne}
    \vskip -0.15in
\end{figure}

\textbf{Evaluation Protocol.}
To ensure our evaluation protocol adheres to high standards of reproducibility and comparability, we used a 10-fold cross-validation strategy, dividing each dataset into three parts: 80\% for training, 10\% for validation, and 10\% for testing. During training, we kept track of the validation accuracy after each epoch. Our early stopping criterion halted training when there was no improvement in validation accuracy for 100 epochs, within a total training span of 300 epochs. The epoch with the highest validation accuracy was considered optimal. Finally, we assessed this optimal performance on the test data, ensuring our results were consistent and reliable.

\begin{table}[t]
    \caption{Graph classification in terms of accuracy (\%) with standard deviation for different graph representation learning methods. The best results are \textbf{boldfaced}. \textbf{OOM}: GPU memory is insufficient to accommodate the requirements.}
    % \vskip 0.1in
    \label{tab_relationships}
    \begin{center}
    \scalebox{0.83}{
    \begin{tabular}{lllllll}
    \toprule
    \multicolumn{1}{@{}c@{}}{\multirow{1}{*}{Dataset}} &
    % \multicolumn{1}{@{}c@{}}{GIN} & 
    \multicolumn{1}{@{}c@{}}{Aug} & 
    \multicolumn{1}{@{}c@{}}{Mixup} & 
    \multicolumn{1}{@{}c@{}}{GraphCL} & 
    \multicolumn{1}{@{}c@{}}{MVGRL} & 
    \multicolumn{1}{@{}c@{}}{Relating-Up} \\   
    \midrule
    MUTAG     
        % & 85.70±9.08 
        & 83.40±11.18 
        & 84.04±9.65
        & 81.32±5.99 
        & 82.12±9.21 
        & \textbf{86.70±7.55} \\
    PTC-FM   
        % & 57.89±4.13 
        & 60.45±5.33  
        & 61.61±4.63 
        & 58.13±4.31 
        & 59.29±6.70 
        & \textbf{64.47±4.06} \\
    PTC-MM   
        % & 62.53±7.91 
        & 63.39±5.83 
        & 64.90±6.10 
        & 65.57±8.42 
        & 63.93±11.2 
        & \textbf{66.38±4.10} \\
    PTC-FR   
        % & 66.07±4.28 
        & 66.79±7.43 
        & 66.64±3.32 
        & 65.42±8.86
        & 66.69±7.64
        & \textbf{68.93±5.56} \\
    PTC-MR   
        % & 57.56±6.29
        & 57.64±6.50  
        & 59.58±5.66 
        & 59.29±7.51 
        & 58.40±8.28 
        & \textbf{62.46±5.80} \\
    COX2     
        % & 83.50±3.63
        & 81.09±3.55 
        & 81.80±4.71 
        & 82.97±4.43 
        & 82.85±4.63 
        & \textbf{83.72±4.10} \\
    COX2-MD  
        % & 60.46±9.03 
        & 60.81±8.08 
        & 61.10±8.15
        & 60.43±6.91 
        & 64.37±9.30 
        & \textbf{67.63±8.40} \\
    PROTEINS  
        % & 73.67±3.57
        & 73.60±4.32 
        & 73.82±4.73
        & 74.02±6.16 
        & 75.11±4.83 
        & \textbf{75.91±4.41} \\
    DD       
        % & 76.57±2.55 
        & 78.18±2.96 
        & 77.76±4.13 
        & 78.32±8.36
        & OOM        
        & \textbf{79.12±3.84} \\
    NCI1      
        % & 79.39±1.42
        & 75.65±2.49 
        & 78.39±1.63 
        & 77.18±2.23 
        & 73.65±3.48 
        & \textbf{80.66±2.65} \\
    NCI109    
        % & 78.41±2.45 
        & 78.84±3.02
        & 78.83±2.19 
        & 78.04±1.26 
        & 73.27±2.31 
        & \textbf{80.63±2.10} \\
    ENZYMES   
        % & 40.00±5.77 
        & 40.33±3.09 
        & 40.33±3.09 
        & 41.83±4.81 
        & 41.33±5.78 
        & \textbf{42.33±5.59} \\
    \midrule
    IMDB-BINARY    
        % & 73.70±2.69
        & 71.90±2.47 
        & 73.80±3.12 
        & 72.90±3.39 
        & 71.80±4.21 
        & \textbf{75.80±3.49} \\
    IMDB-MULTI    
        % & 49.40±3.39 
        & 51.47±3.21 
        & 50.80±3.08 
        & 50.33±2.24 
        & 49.20±3.99 
        & \textbf{52.00±3.35} \\
    REDDIT-BINARY 
        % & 72.70±3.71 
        & 71.60±3.48 
        & 71.45±2.16 
        & 73.30±2.62 
        & 77.40±3.62        
        & \textbf{78.90±2.77} \\
    REDDIT-MULTI-5K
        % & 51.23±2.71 
        & 46.73±1.84 
        & 45.91±4.00
        & 51.63±3.15 
        & OOM       
        & \textbf{54.11±2.10} \\
    \bottomrule
    \end{tabular}}
    \end{center}
    \vskip -0.1in
\end{table}

\subsection{Performance comparison with GNN backbones}
\textbf{GNN Backbones.} In keeping with the focus of our research on enhancing GNN architectures to leverage inter-graph relationships, we primarily compare our module against widely used GNN backbones. Specifically, we assess the performance improvements achieved by integrating the Relating-Up module with various GNN backbones. The comparative results, as depicted in Table \ref{tab_backbone}, illustrate both the original performance of the GNN backbones and the enhancements gained through Relating-Up integration. The most outstanding results within each group are highlighted in bold. Our experiments reveal that Relating-Up consistently boosts the performance of GNN backbones across diverse datasets. Additionally, we have also evaluated Relating-Up using the OGBG-MOLHIV dataset. As shown in Table \ref{tab:ogb}, the results demonstrate significant improvements in ROC-AUC scores with the integration of Relating-Up. These improvements underscore the efficacy of the Relating-Up module in enhancing the predictive performance of GNN backbones on a challenging dataset."

% Specifically, in bioinformatics domain datasets, our approach registers an average performance increase of 2.75±2.08\% across different backbones. This increase breaks down as follows: 2.28±1.84\% for GraphSAGE, 2.69±2.35\% for GCN, and 3.28±2.17\% for GIN. Similarly, in social domain datasets, we observe an average improvement of 1.95±1.40\%. This includes enhancements of 0.62±0.37\% for GraphSAGE, 2.03±1.19\% for GCN, and 3.20±1.38\% for GIN.

\textbf{Empirical Exploration.} As shown in Figure \ref{fig_heatmap} and \ref{fig_tsne}, the embedding distribution learned by the Relating-Up is more compact in the same class and more separated in different classes. This enhancement results from the ability to link and integrate relationships across graphs, an aspect often neglect in standard GNN architectures. Overall, these results reinforce the effectiveness of Relating-Up module and highlight the versatility and robustness of Relating-Up as a valuable enhancement to graph representation learning.

\textbf{Advanced Pooling Methods.} Additionally, as presented in Table \ref{tab_pooling}, Relating-Up demonstrates consistent improvements across various datasets with advanced pooling methods. In the bioinformatics domain, the average performance boost is 2.53±1.57\% across all pooling methods, with individual improvements of 2.66±1.42\% for DiffPool, 2.04±2.05\% for GAPool, and 0.90±1.20\% for SAGPool. This highlights its potential to significantly impact graph pooling scenarios, with a slightly more pronounced effect on pooling methods compared to the underlying GNN backbones. In social domain datasets, the average performance increase across pooling methods is 1.83±1.04\%, with individual improvements of 2.2±0.68\% for DiffPool, 1.69±1.58\% for GAPool, and 1.59±1.05\% for SAGPool.

\textbf{Other Competitive Methods.} The experimental results detailed in \Cref{tab:adv} demonstrate that Relating-Up exhibits competitive performance compared to other methods that claim to incorporate inter-graph relationships.

\textbf{Hyperparameter Sensitivity.} To balance the feedback training strategy, three hyper-parameters $\alpha$, $\beta$, and $T$ are introduced. Through the experiments detailed in \Cref{sec:hyper}, we find out that these hyperparameters have impacts on the performance.

\textbf{Batch Size.} We examines the Relating-Up response to varying batch size in GNN architectures, as detailed in Table \ref{tab_batch}. The results indicate Relating-Up are not significantly hindered by the variations in batch size.

\begin{table}[t]
\vskip -0.1in
\centering
\caption{Graph classification in terms of accuracy (\%) with standard deviation for competitive methods. The best results are \textbf{boldfaced}.}
\vskip 0.1in
\label{tab:adv}
\scalebox{0.9}{
\begin{tabular}{clcccc}
\toprule
Backbone & Method  & MUTAG         & PROTEINS & DD & NCI1 \\
\midrule
\multirow{2}{*}{GCN} & CARE       & 79.80±12.89         & 71.97±5.65          & 75.97±4.54          & \textbf{79.39±2.03} \\
                     & Relating-Up & \textbf{86.20±7.87} & \textbf{74.84±3.78} & \textbf{77.84±3.30} & 77.19±1.65 \\
\midrule
\midrule
\multirow{2}{*}{GIN} & CARE        & 85.18±8.36          & 72.51±6.89          & 74.79±4.10          & 81.30±1.14 \\
                     & Relating-Up & \textbf{86.70±7.55} & \textbf{75.91±4.40} & \textbf{79.12±3.84} & \textbf{80.66±2.65} \\
\bottomrule
\end{tabular}}
\vskip 0.1in
\end{table}

\begin{table}[t]
\caption{Ablation study results with average accuracy (\%) and standard deviation. The best results are \textbf{boldfaced}.}
\label{tab_ablation}
\begin{center}
\scalebox{0.7}{
\begin{tabular}{cccccccccc}
    \toprule
    & $\mathcal{L}_{distill}$ & $\mathcal{L}_{hint}$ & MUTAG & PTC-FM  & COX2 & COX2-MD & PROTEINS & DD & NCI1 \\
    \midrule
    \textbf{A1} & \ding{56} & \ding{56} & 82.06±9.39 & 58.45±7.46 & 77.95±0.90  & 51.15±1.30 & 68.85±6.65 & 70.35±9.22 & 72.78±8.13 \\
    \textbf{A2} & \ding{56} & \ding{52} & 85.12±9.70 & 60.14±5.29 & 82.23±3.14  & 66.42±8.85 & 75.20±2.76 & 77.96±3.86 & 79.51±1.77 \\
    \textbf{A3} & \ding{52} & \ding{56} & 80.88±11.05 & 63.61±4.09 & 80.07±4.24 & 52.82±7.35 & 74.30±3.10 & 75.64±4.18 & 77.13±6.82 \\
    
    \midrule
    \midrule
    
    Full Model & \ding{52} & \ding{52} & \textbf{86.70±7.55} & \textbf{64.47±4.06} & \textbf{83.72±4.10}  & \textbf{67.63±8.40} & \textbf{75.91±4.41} & \textbf{79.12±3.84} & \textbf{80.66±2.65} \\
    \bottomrule
\end{tabular}}
\end{center}
\vskip -0.1in
\end{table}

\textbf{Time Efficiency.} Our computational efficiency study detailed in \Cref{sec:time} reveal that while our module introduces additional computational steps, the overall increase in processing time is marginal compared to the significant performance gains achieved. 

\subsection{Performance comparison with other graph representation learning paradigms}
To further evaluate the effectiveness of proposed Relating-Up, we compare it against other graph representation learning paradigms. The comparison includes GraphAug, Mixup for graph, GraphCL, and MVGRL. These methods represent a diverse set of strategies for enhancing graph representation learning. The experimental results are shown in the Table \ref{tab_relationships}. It can been seen from the experiments results that Relating-Up consistently outperforms these methods that require an predefined relationships. We attributed this improvement to the dynamic understanding inter-graph relationships of Relating-Up.

\subsection{Ablation Studies}
The feedback training strategy entails dual-focused training: firstly on the original graph representations, and secondly on the relation-aware representations distilled from inter-graph relationships. Our experimental setup included three variations: (\textbf{A1}) a GNN module implementing Relating-Up without feedback training strategy;  A GNN model with Relating-Up but (\textbf{A2}) remove $\mathcal{L}_{distill}$, and (\textbf{A3}) remove $\mathcal{L}_{hint}$. The experimental results, detailed in Table \ref{tab_ablation}, showed an intriguing trend. The standalone implementation of relation-aware representation (\textbf{A1}) did not surpass baseline modules in performance, which emphasizes the importance of the feedback training strategy in the Relating-Up framework. The results demonstrate that effective integration and utilization of insights from inter-graph relationships hinge on a complete feedback training strategy. 

\section{Conclusion}

In this work, we introduced Relating-Up, an innovative module designed to enhance GNNs by enabling them to dynamically explore and leverage inter-graph relationships. Integrating a relation-aware encoder with a feedback training strategy, Relating-Up transforms GNNs beyond their standard intra-graph focus, allowing them to capture complex dynamics across various graphs. This module, adaptable to a wide range of existing GNN architectures, significantly boosts their performance across diverse benchmark datasets. By providing a deeper, more nuanced understanding of graph data, Relating-Up marks a significant advancement in graph representation learning.

\textbf{Limitation.} While Relating-Up presents significant benefits, it may require some adaptation to seamlessly integrate with unsupervised graph representation learning paradigms, especially those based on contrastive learning. However, we believe that with further research and development, this limitation can be addressed, making it more readily extendable to unsupervised learning paradigms.

\bibliographystyle{unsrt}
{\small
\bibliography{ref}}

\newpage
\appendix
\renewcommand\thetable{S\arabic{table}}
\renewcommand\thefigure{S\arabic{figure}}
\setcounter{table}{0}
\setcounter{figure}{0}

\section{Algorithm of feedback training strategy}
Our proposed feedback training strategy is summarized in \Cref{alg:feedback}.

\begin{algorithm}[!ht]
   \caption{Relating-Up Feedback Training Strategy}
   \label{alg:feedback}
\begin{algorithmic}
   \STATE {\bfseries Input:} a set of training graphs $\mathbb{G}$ with label $\mathbb{Y}$
   \STATE {\bfseries Initialization:} Graph Encoder $f_{\mathcal{G}}$, Relation Encoder $f_{\mathcal{R}}$, Classifiers $f^{\mathcal{G}}_{\theta}$ and $f^{\mathcal{R}}_{\theta}$
   \STATE {\bfseries Output:} Refined Graph Encoder $f_{\mathcal{G}}$

   \REPEAT
        \STATE // Compute Original/Refined Graph Representation
        \STATE$\mathbf{E} \leftarrow f_{\mathcal{G}}(\mathbb{G})$
        
        \STATE // Compute Relation-Aware Representations
        \STATE $\mathbf{R} \leftarrow f_{\mathcal{R}}(\mathbf{E})$ 
        % \STATE // Loss Computation for Feedback Training
        \STATE // CrossEntropy loss
        \STATE $\mathcal{L}_{class} \leftarrow \text{CrossEntropy}(f^{\mathcal{G}}_{\theta}(\mathbf{E}),\mathbb{Y}) + $
        \STATE\hspace{38.5pt} $\text{CrossEntropy}(f^{\mathcal{R}}_{\theta}(\mathbf{R}),\mathbb{Y})$
        \STATE // KL Divergence loss
        \STATE $\mathcal{L}_{distill} \leftarrow D_{kl}(f^{\mathcal{G}}_{\theta}(\mathbf{E}), f^{\mathcal{R}}_{\theta}(\mathbf{R}))$
        \STATE // L2 loss from hints
        \STATE $\mathcal{L}_{hint} \leftarrow \parallel \mathbf{R} - \mathbf{E} \parallel_2^2$
        \STATE // Total loss 
        \STATE $\mathcal{L} \leftarrow \alpha \mathcal{L}_{class} + (1 - \alpha) \mathcal{L}_{distill} + \beta \mathcal{L}_{hint}$

        \STATE Backpropagate $\mathcal{L}$
        \STATE Update parameters of $f_{\mathcal{G}}$, $f_{\mathcal{R}}$, $f^{\mathcal{G}}_{\theta}$, and $f^{\mathcal{R}}_{\theta}$
        
    \UNTIL{Convergence or reaching maximum training times}
\end{algorithmic}
\end{algorithm}

\section{Detailed Experiment Settings}\label{sec:sec1}
\subsection{Datasets}

\begin{table}[h]
\caption{Data statistics.}
\vskip 0.1in
\label{tabs1}
\begin{center}
\scalebox{0.8}{
    \begin{tabular}{llccccc}
        \toprule
        \multicolumn{2}{c}{Dataset} & \#Graph & \#Classes & \#Nodes & \#Edges & \#Features \\
        \midrule
        \multirow{11}{*}{\rotatebox{90}{Bio.}}
        & MUTAG           & 188  & 2 & 17.93  & 19.97  & 7  \\
        & PTC-FM         & 349  & 2 & 14.11  & 14.48  & 18 \\
        & PTC-MM         & 351  & 2 & 14.56  & 15.00  & 19 \\
        & PTC-FR         & 336  & 2 & 13.97  & 14.32  & 20 \\
        & PTC-MR         & 344  & 2 & 14.29  & 14.69  & 18 \\
        & COX2-MD        & 303  & 2 & 26.28  & 335.12 & 7  \\
        & COX2            & 467  & 2 & 41.22  & 43.45  & 35 \\
        & PROTEINS        & 1113 & 2 & 39.06  & 72.82  & 3  \\
        & DD              & 1178 & 2 & 284.32 & 715.66 & 89 \\
        & NCI1            & 4110 & 2 & 29.87  & 32.30  & 37 \\
        & NCI109          & 4127 & 2 & 29.68  & 32.13  & 38 \\
        & ENZYMES         & 600  & 6 & 32.63  & 64.14  & 3  \\
        \midrule
        \multirow{4}{*}{\rotatebox{90}{Social.}}
        & IMDB-BINARY     & 1000 & 2 & 19.77  & 96.53  & - \\
        & IMDB-MULTI      & 1500 & 3 & 13.00  & 69.54  & - \\
        & REDDIT-BINARY   & 2000 & 2 & 429.63 & 497.75 & - \\
        & REDDIT-MULTI-5K & 4999 & 5 & 508.52 & 594.87 & - \\
    \bottomrule
    \end{tabular}}
\end{center}
\end{table}

All dataset are publicly available and represent a relevant subset of those most frequently used in literature for GNNs comparison. Specifically, this collection includes 16 datasets collected from the TUDataset \cite{KKMMN2016} and the OGBG-MOLHIV dataset collected from the Open Graph Benchmark (OGB) \cite{hu2021open} . The bioinformatics domain including MUTAG, PTC-FM, PTC-MM, PTC-FR, PTC-FM, COX2, COX2-MD, PROTEINS, DD, NCI1, NCI109, and ENZYMES. The  social networks datasets including IMDB-BINARY, IMDB-MULTI and REDDIT-BINARY, REDDIT-MULTI-5K. The addition of OGBG-MOLHIV further enriches our experiments by incorporating a more challenge modern dataset. Notably, since the node features are not present for social domain datasets, we assign a uniform uninformative feature to all nodes for REDDIT datasets, and we we utilize one-hot encoding of node degrees for IMDB datasets \cite{errica2019fair}. Detailed statistics and properties of these datasets are presented in Table \ref{tabs1}. 

For all datasets, we employed a 10-fold cross-validation strategy for all datasets, dividing each dataset into training, validation, and test subsets in a 8:1:1 ratio, respectively. All experiments were conducted using the exact same fold index to ensure fair comparisons.

\subsection{Model Architectures Hyperparameters}
The hyperparameters of GNN backbone is configured with five GNN layers, each comprising 128 hidden units. Unless otherwise stated, we opted for a global sum pooling after GNNs to obtain graph representations. For graph augmentation, we random choice 2 compose from subgraphs induced by random walks (RWS), node dropping, feature masking, and edge removing. For graph contrastive learning model, i.e., GraphCL and MVGRL, the hyperparameters of these models are taken from the original official examples. For Relating-Up, we do a hyperparameter search of $\alpha$ from $[0.1, 0.2, 0.3, 0.4]$, $\beta$ from $[1e-6, 1e-5, 1e-4]$, and $T$ from $[2, 3, 4]$. All the codes were implemented using PyTorch and PyTorch Geometric packages. The experiments were conducted in a Linux server with Intel(R) Core(TM) i9-13900KF CPU (3.0GHz), GeForce GTX 4090 GPU, and 64GB RAM.

\section{Hyperparameter Analysis}

\begin{figure}[!t]
\begin{center}
\includegraphics[width=0.9\linewidth]{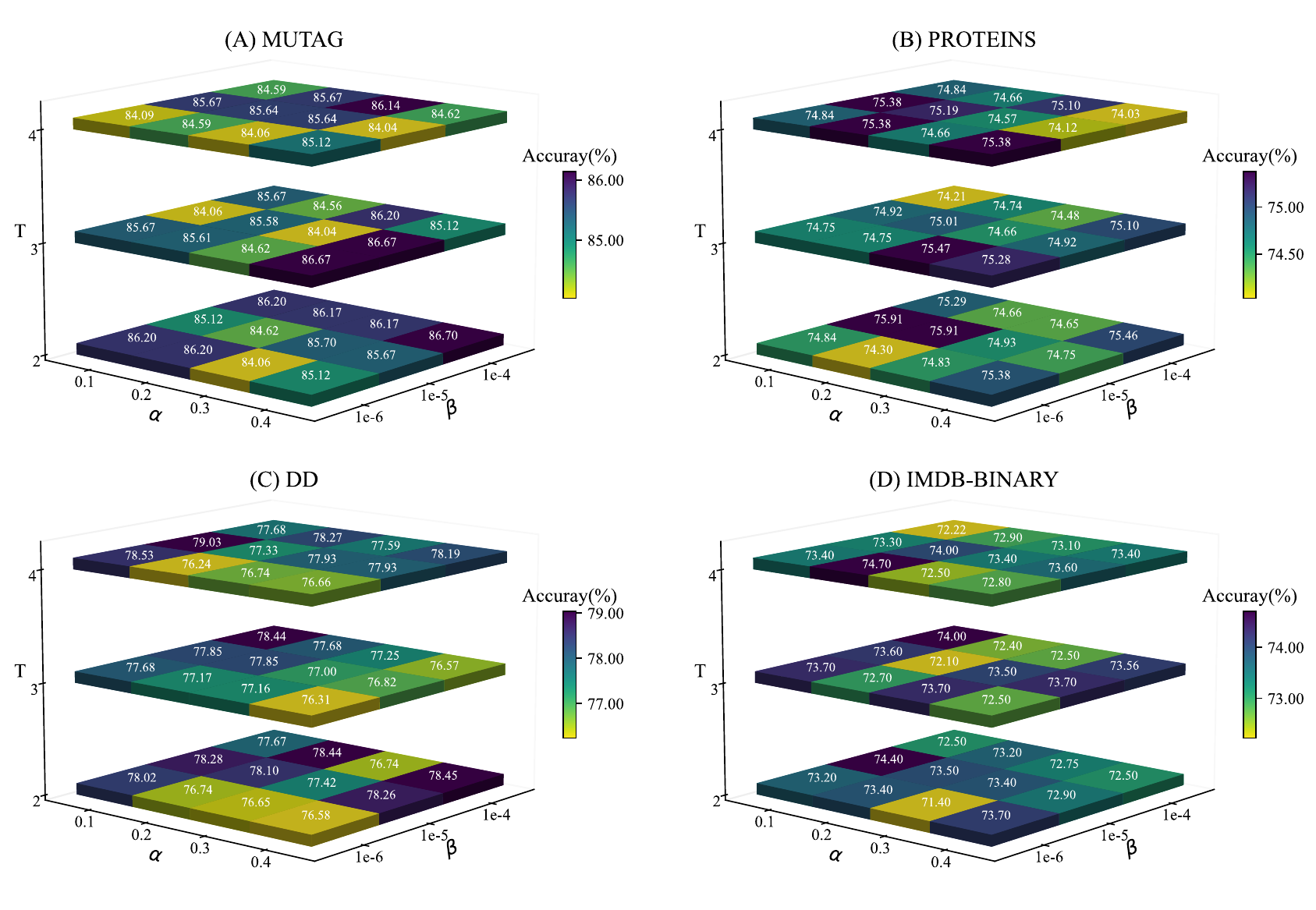}
\vskip -0.1in
\caption{Results of hyperparameter analysis.}
\label{fig:hparams}
\end{center}
\end{figure}

\label{sec:hyper}
In this section, we present a sensitivity analysis of hyperparameters for the proposed Relating-Up across four distinct dataset: MUTAG, PROTEINS, DD, and IMDB-BINARY. In our experimental setup, Relating-Up integrated with a 5 layers GIN backbone with each layer comprising 128 hidden units. We focus on three hyperparameters introduced by the feedback training strategy: $\alpha$ (balancing cross-entropy loss and KL divergence, varied within the range of $[0.1, 0.2, 0.3, 0.4]$), $\beta$ (L2 hint loss, tested across values in the specturm of $[1e-6, 1e-5, 1e-4]$), and $T$ (the temperature parameter for KL divergence, experimented with in the range of $[2, 3, 4]$). The experimental results are shown in Figure \ref{fig:hparams}.

\subsection{Analysis of $\alpha$}
$\alpha$ serves as the balance between the cross-entropy loss for labeled graph entities and the KL divergence. It was observed that the optimal value of $\alpha$ is highly dataset-specific. Lower values of $\alpha$ tend to give more weight to the KL divergence, thus encouraging the model to leverage relation-aware representations more effectively. In contrast, higher values of $\alpha$ emphasize the cross-entropy loss, focusing the model on learning from the original graph representations. The optimal value of $\alpha$ depends on the specific characteristics of the dataset and the complexity of the graph structures involved.

\subsection{Analysis of $\beta$}
The role of $\beta$ is to manage the representation hints loss, aiming to reduce the gap between the relation-aware and the original graph representations. Altering the value of $\beta$ has a significant impact on how the model aligns these two representation spaces. Higher values of $\beta$ enforce a stronger congruence, potentially leading to a more harmonious integration of knowledge from inter-graph relationships. However, too high a $\beta$ can result in overfitting to the relation-aware representations, compromising the generalization abilities.

\subsection{Analysis of $T$}
The temperature parameter $T$ controls the softening of the probability distribution across classes. A higher $T$ leads to a softer distribution, which can be beneficial in scenarios where the inter-graph relationships are not entirely reliable or when a more generalized learning approach is desired.

\subsection{Dataset-Specific Optimal Settings}
Across all datasets, the results underscore the need for dataset-specific hyperparameter tuning in the Relating-Up module. The optimal settings for $\alpha$, $\beta$, and $T$ varied notably across different datasets, reflecting the diverse nature of graph structures, complexity, and label distributions in graph-based learning tasks. This highlights the importance of a tailored approach in hyperparameter optimization for enhancing the performance of graph neural networks in various applications.

\section{Time Efficiency}\label{sec:time}
The Relating-Up module incorporation of a relation encoder to process relationships across graph and employs a dynamic feedback training strategy to refine inter-graph relationships. Although incorporating these components introduce computational overhead during training, the exclusion of these components during the inference phase ensures that the time efficiency is comparable to conventional GNN models. To investigate the time efficiency of the Relating-Up module, we compared the runtime against standard GNN backbone. We conducted experiments on several representative datasets, measuring the time taken for training and inference per epoch. 

\subsection{Training Time}
The training time of Relating-Up is expected to be longer than that of the baseline GNN models due to the additional components. Specifically, the relation encoder processes the set of all graph representations in a batch. This operation adds complexity and increases training time. Moreover, the feed learning strategy, also contributes to the longer training times as it requires additional forward and backward passes during the optimization process. Table \ref{tab_time} provides a comprehensive overview of the training times for various methods, including Relating-Up, on different datasets. The relative training time (compared to the GNN backbone) is \textbf{boldfaced}, while the actual time in seconds is \textit{italicized}. As observed, Relating-Up exhibits a slight increase in training time compared to the original GNN backbones but remains significantly more efficient than other complex methods like MVGRL and GraphCL.

\begin{table}[!ht]
\caption{Training time per epoch for various methods. The relative training time compared to the backbone is \textbf{boldfaced}, and the actual time in seconds is \textit{italicized}.}
\label{tab_time}
\begin{center}
\vskip 0.1in
    \scalebox{0.9}{
    \begin{tabular}{lcccccccc}
    \toprule
     & Methods & MUTAG & PROTEINS & DD & IMDB-BINARY \\
    \midrule
    \multirow{8}{*}{\rotatebox{90}{GCN}}
                    & \multirow{2}{*}{Original}
                    & \textbf{1} & \textbf{1} & \textbf{1} & \textbf{1} \\
                    && \textit{0.036} & \textit{0.128} & \textit{0.815} & \textit{0.134} \\
                    \cmidrule(l){3-6}
                    
                    & \multirow{2}{*}{MVGRL}
                    & \textbf{6.08} & \textbf{15.01} & \textbf{49.58} & \textbf{5.85} \\
                    && \textit{0.216} & \textit{1.924} & \textit{40.402} & \textit{0.782} \\
                    \cmidrule(l){3-6}
                    
                    & \multirow{2}{*}{GraphCL}
                    & \textbf{2.0} & \textbf{2.0} & \textbf{2.2} & \textbf{1.61} \\
                    && \textit{0.071} & \textit{0.256} & \textit{1.791} & \textit{0.215} \\
                    \cmidrule(l){3-6}
                    
                    & \multirow{2}{*}{Relating-Up} 
                    & \textbf{1.22} & \textbf{1.15} & \textbf{1.02} & \textbf{1.12} \\
                    && \textit{0.043} & \textit{0.148} & \textit{0.83} & \textit{0.15} \\
    \midrule
    \multirow{8}{*}{\rotatebox{90}{GIN}}
                    & \multirow{2}{*}{Original}
                    & \textbf{1} & \textbf{1} & \textbf{1} & \textbf{1} \\
                    && \textit{0.036} & \textit{0.128} & \textit{0.815} & \textit{0.134} \\
                    \cmidrule(l){3-6}
                    
                    & \multirow{2}{*}{MVGRL}
                    & \textbf{6.57} & \textbf{14.73} & \textbf{55.48} & \textbf{9.0} \\
                    && \textit{0.233} & \textit{1.888} & \textit{45.216} & \textit{1.203} \\
                    \cmidrule(l){3-6}
                    
                    & \multirow{2}{*}{GraphCL}
                    & \textbf{2.07} & \textbf{1.99} & \textbf{2.11} & \textbf{2.47} \\
                    && \textit{0.074} & \textit{0.256} & \textit{1.719} & \textit{0.331} \\
                    \cmidrule(l){3-6}
                    
                    & \multirow{2}{*}{Relating-Up} 
                    & \textbf{1.22} & \textbf{1.15} & \textbf{1.02} & \textbf{1.12} \\
                    && \textit{0.043} & \textit{0.148} & \textit{0.83} & \textit{0.15} \\
    \bottomrule
    \end{tabular}}
\end{center}
\end{table}

\subsection{Inference Time}
\begin{figure}[!t]
    \begin{center}
        \centerline{\includegraphics[width=0.9\linewidth]{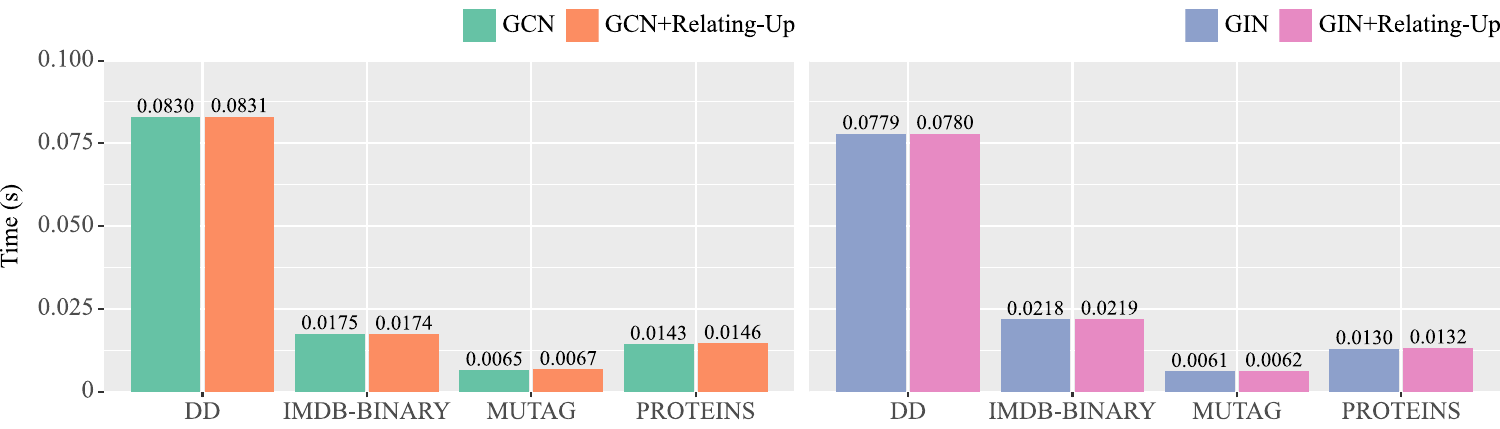}}
        \caption{Inference time (in seconds) of backbone and Relating-Up.}
        \label{inference_times}
    \end{center}
    \vskip -0.2in
\end{figure}

As shown in Figure \ref{inference_times}, during inference, the Relating-Up module operates more efficiently compared to the training phase. Since the relation encoder and the additional components for feedback training strategy are not utilized during inference, the time efficiency is comparable to the standard GNN models. This optimization ensures that while Relating-Up adds computational overhead during training, it does not significantly impact the performance during the inference phase, which is crucial for practical applications.

In summary, while Relating-Up introduces additional computational costs during training, it maintains efficient inference times. The extra time taken during training is a trade-off for the enhanced performance and the ability to leverage inter-graph relationships, which can be crucial for complex graph analysis tasks. The efficiency during inference makes Relating-Up a practical solution for real-world applications where quick response times are essential.

\section{Experimentation on Batch Size}

\begin{table*}[!t]
\caption{Experimentation of batch size with GCN as backbone.}
\label{tab_batch}
\vskip 0.1in
\begin{center}
\begin{tabular}{lcccc}
\toprule

Dataset                      & Methods     & Batch size = 16      & Batch size = 32      & Batch size = 128 \\
\midrule
\multirow{4}{*}{MUTAG}       & GCN         & 82.49±11.31          & 86.20±10.31          & 84.09±10.23 \\
                             & Relating-Up & \textbf{86.14±8.75}  & \textbf{86.75±7.11}  & \textbf{86.20±7.87}  \\
                             \cmidrule(l){2-5}
                             & GIN         & 84.04±6.29           & 81.37±11.17          & 85.70±9.08 \\
                             & Relating-Up & \textbf{86.20±7.14}  & \textbf{86.20±8.51}  & \textbf{86.70±7.55}  \\
\cmidrule(l){1-5}
\multirow{4}{*}{PROTEINS}    & GCN         & 73.66±3.41           & 74.56±3.25           & 73.49±3.71  \\
                             & Relating-Up & \textbf{74.84±2.66}  & \textbf{75.11±3.50}  & \textbf{74.84±3.78}  \\
                             \cmidrule(l){2-5}
                             & GIN         & 74.39±3.33           & 73.49±3.56           & 73.67±3.57 \\
                             & Relating-Up & \textbf{74.84±2.76}  & \textbf{75.11±2.17}  & \textbf{75.91±3.41}  \\
\cmidrule(l){1-5}
\multirow{4}{*}{DD}          & GCN         & 77.25±2.07           & 76.07±3.56           & 76.66±3.50  \\
                             & Relating-Up & \textbf{78.52±3.13}  & \textbf{78.27±3.80}  & \textbf{78.02±2.02}  \\
                             \cmidrule(l){2-5}
                             & GIN         & 76.31±2.44           & 76.32±3.31           & 76.57±2.55  \\
                             & Relating-Up & \textbf{78.44±2.96}  & \textbf{78.45±2.77}  & \textbf{79.12±3.84}  \\
\cmidrule(l){1-5}
\multirow{4}{*}{IMDB-B}      & GCN         & 74.20±2.89           & 72.80±3.68           & 73.00±3.41  \\
                             & Relating-Up & \textbf{75.38±2.44}  & \textbf{75.91±2.57}  & \textbf{74.50±3.35}  \\
                             \cmidrule(l){2-5}
                             & GIN         & 72.40±3.98           & 73.70±3.26           & 73.70±2.69   \\
                             & Relating-Up & \textbf{74.70±2.28}  & \textbf{74.40±2.80}  & \textbf{75.80±3.49}  \\
\bottomrule
\end{tabular}
\end{center}
\end{table*}

The experimentation on batch size reveals how the Relating-Up module responds to different batch sizes when integrated with GNN architectures. The study covers batch sizes of 16, 32, and 128, providing a broad spectrum of sizes from small to large. This range allows for an in-depth analysis of the impact of batch size on the learning dynamics and performance of the Relating-Up-enhanced models. The results, presented in the table \ref{tab_batch}, illustrate the performance on various datasets, comparing the baseline GCN with the Relating-Up enhanced version across different batch sizes.

Notably, the Relating-Up module consistently outperforms the baseline GCN model across all batch sizes for each dataset. This indicates that the effectiveness of Relating-Up is not heavily dependent on the batch size, showcasing its robustness and adaptability. The ability of Relating-Up to enhance graph representation learning is evident across different batch configurations, suggesting that it can be effectively integrated into various GNN architectures without the need for extensive hyperparameter tuning related to batch size.

Furthermore, the relatively stable performance of Relating-Up across different batch sizes also indicates that the mechanisms for leveraging inter-graph relationships and incorporating feedback training strategies are not significantly hindered by the variations in batch size. This aspect is particularly important for practical applications where batch size may vary based on the available computational resources or specific requirements of the task.

\end{document}